\documentclass[journal]{IEEEtran}

\usepackage{amsmath,amsfonts,bm}
\usepackage{graphicx, subcaption, tikz}
\usepackage{booktabs, multirow}
\usepackage[hidelinks]{hyperref}
\usepackage{url}
\newcommand{\res}[2]{#1 \scriptsize (#2)}
\newcommand{\rpm}[2]{#1 \scriptsize $\pm$ #2}
\newcommand{\pca}{Rec-PCA}
\newcommand{\vect}[1]{\bm{\MakeLowercase{#1}}}

\title{Disjoint or Overlapping? Inference Windowing for Reconstruction-Based Time Series Anomaly Detection}

\author{
Guillaume Coulaud,
Reza Akbarinia,
Florent Masseglia
\thanks{G. Coulaud is with the University of Montpellier, Inria, CNRS, LIRMM, Montpellier, France (email: guillaume.coulaud@inria.fr).}
\thanks{R. Akbarinia is with Inria, University of Montpellier, CNRS, LIRMM, Montpellier, France (email: reza.akbarinia@inria.fr).}
\thanks{F. Masseglia is with Inria, University of Montpellier, CNRS, LIRMM, Montpellier, France (email: florent.masseglia@inria.fr).}
}

\begin{document}

\maketitle

\begin{abstract}
Reconstruction-based methods are widely used for time series anomaly detection, where models are trained to reconstruct subsequences, and anomalies are identified through reconstruction errors. However, reported results are often hard to compare due to heterogeneous evaluation practices and under-specified inference procedures.

In this paper, we revisit reconstruction-based anomaly detection in the univariate offline setting and study the role of the inference stride, which controls whether subsequences are processed as disjoint windows or with overlap. We propose a unified training, tuning, and multi-seed evaluation protocol on the curated TSB-AD benchmark, and study how overlapping inference affects anomaly detection performance for a range of reconstruction models, including PCA-based baselines, DLinear, an AutoEncoder, TimesNet, and Transformer variants. The results show that across all models, overlapping windows yield consistent improvements, with average relative gain up to +28\%, and can alter method rankings. We further analyze variability across datasets, random seeds, and hyperparameter configurations. Finally, we complement the benchmark study with an evaluation on the full UCR archive using localization criteria aligned with sliding-window reconstruction. Overall, our results highlight that reconstruction-based anomaly detection performance depends not only on model architecture and training, but also on inference choices, motivating a clear and reproducible protocol. Our results show that reconstruction-based baselines achieve strong performance on both TSB-AD and UCR benchmarks, supporting them as competitive and practical approaches for univariate time series anomaly detection.
\end{abstract}


\section{Introduction}

Time series anomaly detection (TSAD) plays a central role in many real-world applications, including industrial monitoring~\cite{yanComprehensiveSurveyDeep2024}, finance~\cite{yangDetectionAnomalyStock2020}, and healthcare~\cite{yangDeepLearningTechnologies2023}. In recent years, reconstruction-based approaches have emerged as a strong paradigm~\cite{zamanzadehdarbanDeepLearningTime2024}, leveraging models such as autoencoder~\cite{bank2023autoencoders}, recurrent neural network~\cite{medsker2001recurrent}, and Transformer~\cite{vaswaniAttentionAllYou2017}.

These methods rely on the assumption that a model trained on normal data can accurately reconstruct regular patterns, while anomalies lead to higher reconstruction errors. Reconstruction-based methods typically operate on fixed-size subsequences (or window) extracted from the original time series. While the training procedure is generally standardized across the literature using sliding windows with a stride (or step) of 1, the inference strategy remains largely understudied. In particular, the inference stride determines whether subsequences are processed using overlapping or disjoint windows, directly affecting how anomaly scores are reconstructed over time.

Recent benchmarks and evaluation frameworks, including TAB~\cite{qiuTABUnifiedBenchmarking2025}, TSB-AD~\cite{liuElephantRoomReliable2025}, and Time Series Library~\cite{wangDeepTimeSeries2024}, consider only disjoint windows during inference, \textit{i.e.}, a stride equal to the window length. This raises an important question: does the choice of inference stride, and more generally the use of overlapping versus disjoint windows, affect the anomaly detection performance of reconstruction-based methods?



In this paper, our goal is to answer this question by assessing  reconstruction-based methods under a controlled protocol. In particular, we study how the windowing strategy at inference affects  anomaly detection performance in the case of univariate time series. 

Our contributions can be summarized as follows.
\begin{itemize}
     \item We conduct a controlled empirical study across a diverse collection of time series datasets (TSB-AD and the UCR archive), and provide both benchmark-level and dataset-level performance insights.
    \item We investigate the effect of windowing strategy at inference, discovering that the use of overlapping windows has a significant impact on anomaly detection performance, with average relative gain up to +28\%.
    \item We show that, under careful tuning and using overlapping windows, simple reconstruction-based methods can achieve state-of-the-art performance on the TSB-AD benchmark.
    \item We study the robustness of reconstruction methods to random initialization and hyperparameter selection, illustrating the robustness issues of deep learning methods.
\end{itemize}

The remainder of the paper is organized as follows. We present the related work in Section~\ref{sec:related}. The problem is described in Section~\ref{sec:background}. The experimental protocol is presented in Section~\ref{sec:setup}. We analyze the results in Section~\ref{sec:results} and provide complementary analysis in Section~\ref{sec:complementary}.
We discuss results and takeaways in Section~\ref{sec:discussion} and conclude in Section \ref{sec:conclusion}.

\section{Related Work}
\label{sec:related}
Anomaly detection in time series has been widely studied across data mining, machine learning, and statistics~\cite{chandola2009anomaly,blazquez2021review,boniol2024dive,anomaly2023zhang}. Existing approaches can be broadly grouped into distance-based, density-based, and prediction-based methods. Distance-based techniques detect anomalies by identifying subsequences that differ significantly from others, such as in the Matrix Profile~\cite{yeh2016matrix}, while density-based methods, such as Isolation Forest~\cite{liuIsolationForest2008}, focus on isolating rare observations. Prediction-based approaches rely on forecasting models, from statistical methods such as ARIMA~\cite{yaacobARIMABasedNetwork2010} to deep learning models such as LSTM~\cite{10.1145/3219819.3219845}, and flag deviations between predicted and observed values.

In recent years, reconstruction-based approaches have emerged as a prominent alternative. Rather than estimating similarity, density, or future values, these methods learn to reconstruct typical patterns and detect anomalies through reconstruction errors. Due to their flexibility and compatibility with deep learning models, they have become a central focus of recent work. In the following, we concentrate on this family of methods.

\paragraph{Reconstruction-based Anomaly Detection}

Reconstruction-based methods are widely used for anomaly detection in time series. Early approaches rely on linear techniques such as Principal Component Analysis (PCA)~\cite{takeishi2019shapley}, while more recent methods leverage deep learning models, including autoencoders, recurrent networks, and convolutional architectures. For example, EncDec-AD~\cite{malhotraLSTMbasedEncoderDecoderMultisensor2016} introduces an LSTM encoder-decoder for sequence reconstruction, while LSTM-VAE~\cite{parkMultimodalAnomalyDetector2018} extends this idea using a variational autoencoder. Other works, such as MSCRED~\cite{zhangDeepNeuralNetwork2019}, combine convolutional and recurrent components to model spatio-temporal dependencies.
TimesNet~\cite{wu2022timesnet} models multi-periodicity by transforming one-dimensional signals into a two-dimensional representation based on frequency components estimated through a Fourier transform, then leverage 2d convolutions through inception blocks to reconstruct the input.

More recently, Transformer-based architectures~\cite{vaswaniAttentionAllYou2017} have gained attention for time series analysis~\cite{wenTransformersTimeSeries2023}. Models such as AutoFormer~\cite{wu2021autoformer}, FEDFormer~\cite{zhou2022fedformer} introduce decomposition-based designs that separate trend and seasonal components, typically using moving average filters. Additionally to the attention and decomposition mechanisms, AutoFormer add an autocorrelation mechanism while FEDFormer add frequency enhanced decomposition. These architectures aim to improve the modeling of long-range dependencies and structured temporal patterns.

In this work, we focus exclusively on reconstruction-based models in order to assess their intrinsic anomaly detection capabilities. However, several extensions have been proposed to enhance reconstruction-based approaches. For instance, USAD~\cite{audibertUSADUnSupervisedAnomaly2020a} incorporates adversarial training to better isolate anomalies, while Anomaly Transformer~\cite{xu2021anomaly} combines reconstruction error with attention-based discrepancy measures. Other recent approaches explore contrastive learning strategies to improve the separation between normal and anomalous patterns~\cite{kimContrastiveTimeSeriesAnomaly2024}.

\paragraph{Anomaly Detection Evaluation}

The evaluation of time series anomaly detection methods is challenging due to the ambiguity of anomalies and the diversity of application settings. Existing protocols can be broadly divided into threshold-based and score-based evaluation.

In threshold-based evaluation, models produce anomaly scores that are converted into binary predictions using a thresholding strategy. Metrics such as precision, recall, and F1-score are then computed, but they are highly sensitive to this choice and may not reflect the intrinsic quality of the scores.

To avoid this dependency, score-based evaluation relies on threshold-independent metrics such as AUC-ROC~\cite{fawcettIntroductionROCAnalysis2006} and AUC-PR~\cite{davisRelationshipPrecisionRecallROC2006}, which assess the ranking quality of anomaly scores across all thresholds. However, these metrics treat each time step independently, ignoring the temporal structure of anomalies, which often occur as contiguous subsequences.

To address this limitation, range-based~\cite{tatbulPrecisionRecallTime2018} metrics introduce a tolerance around annotated anomalies by allowing partial scores within a buffer region. Extending this idea, the Volume Under the Surface (VUS)~\cite{boniolVUSEffectiveEfficient2025} aggregates performance over multiple buffer sizes, providing a more comprehensive evaluation across different tolerance levels at an increased computational cost.

\section{Background}
\label{sec:background}
\subsection{Reconstruction Formulation}

Let $\vect{x} = \{x_{t_0}, x_{t_1}, \ldots, x_{t_{n-1}}\}$ denote a univariate time series of length $n$, where $x_{t_i}$ is the observation at time $t_i$. The goal is to assign an anomaly score $s_{t_i}$ to each time step.

Reconstruction-based approaches learn a model $f_\theta$ that reconstructs subsequences extracted from the time series. Given a window size $w$, a subsequence starting at index $i$ is defined as
\begin{equation*}
\vect{x}_{i,w} = \{x_{t_i}, x_{t_{i+1}}, \ldots, x_{t_{i+w-1}}\}.
\end{equation*}

The model produces a reconstruction of the same dimension:
\begin{equation*}
\hat{\vect{x}}_{i,w} = f_\theta(\vect{x}_{i,w}).
\end{equation*}

Let $s$ denote the inference stride. We extract windows of length $w$ starting at indices
\begin{equation*}
\mathcal{J}_s = \{\, ks \mid k \in \mathbb{N},\; 0 \le ks \le n-w \,\}.
\end{equation*}
An index $i$ is covered by the window starting at $j$ if $j \le i \le j+w-1$. We therefore define the set of window start indices whose windows contain $i$ as
\begin{equation*}
\mathcal{I}_i = \{\, j \in \mathcal{J}_s \mid j \le i \le j+w-1 \,\}.
\end{equation*}

For each $j \in \mathcal{I}_i$, let $\hat{x}_{t_i}^{(j)}$ denote the reconstruction of $x_{t_i}$ produced by the model when processing the window $\vect{x}_{j,w}$ (i.e., the value at the corresponding position in $\hat{\vect{x}}_{j,w}$). The reconstruction error associated with this window is
\begin{equation*}
e_{t_i}^{(j)} = \bigl(x_{t_i} - \hat{x}_{t_i}^{(j)}\bigr)^2.
\end{equation*}

Finally, the anomaly score at time $t_i$ is obtained by averaging the reconstruction errors over all windows that contain $i$:
\begin{equation*}
s_{t_i} = \frac{1}{|\mathcal{I}_i|} \sum_{j \in \mathcal{I}_i} e_{t_i}^{(j)}.
\end{equation*}

This formulation supports both overlapping and non-overlapping inference strategies. When the stride is set to $s = w$, subsequences are disjoint, and each observation is evaluated exactly once. When $s < w$, subsequences overlap, and each observation is reconstructed multiple times under different local temporal contexts. Averaging the reconstruction errors reduces sensitivity to the position of the observation within a window and yields a more robust anomaly score.

Let $m$ denote the number of extracted subsequences, where $m = \mathcal{O}(n/s)$. Each subsequence of length $w$ is processed once by the reconstruction model $f_\theta$, yielding a computational cost proportional to $w$ per subsequence. The total inference cost is therefore
\(\mathcal{O}(m w) = \mathcal{O}\left(\frac{n w}{s}\right)\).

In the non-overlapping case ($s = w$), the complexity reduces to $\mathcal{O}(n)$ up to a factor linear in the window size. In the fully overlapping case ($s = 1$), the cost becomes $\mathcal{O}(n w)$ due to maximal reuse of overlapping windows.

\subsection{Evaluation Pipeline}

The anomaly detection process can be decomposed into three stages: training, inference, and scoring. While training procedures are typically well specified, the inference stage is often treated as a fixed component of the pipeline. However, the anomaly score depends not only on the model but also on the inference parameters, in particular the stride $s$. This makes inference a critical component of the evaluation process.

\subsubsection{Training}

During training, the model learns to reconstruct subsequences extracted from the time series. Given a window size $w$, the objective is to minimize a reconstruction loss, typically the mean squared error (MSE), over the training data. During training, the stride is always set to \(s=1\) (maximal overlap). The full training procedure, including hyperparameter optimization and model selection, is described in Section~\ref{sec:setup}.

\subsubsection{Inference}

At inference time, the trained model is applied to subsequences extracted from the time series using a given stride $s$. Reconstruction errors are computed for each subsequence and then concatenated to produce the final anomaly scores.

In practice, most existing works adopt the disjoint setting, corresponding to $s = w$, where each point is evaluated only once. In this work, we explicitly consider both disjoint and overlapping strategies, and systematically analyze their impact on performance. Unless stated otherwise, we always set the stride to \(s=1\) for the overlapping inference.

\subsubsection{Scoring}

Once anomaly scores are obtained for each time step, performance is evaluated using standard metrics for anomaly detection using labeled data. 

\section{Experimental protocol}
\label{sec:setup}

\subsection{Datasets}

We conducted the experiments using the TSB-AD benchmark \cite{liuElephantRoomReliable2025}, a carefully curated aggregation of existing time series anomaly detection datasets designed to address common issues in prior benchmarks. The curation process includes the identification and removal of time series with labeling errors, dataset biases, or insufficient contextual information for effective anomaly detection.  

For the main evaluation, we use the originally proposed subsampled version of the benchmark, consisting of 350 univariate time series selected from the 870 available. This subsampling ensures a more balanced representation across source datasets, preventing larger datasets from disproportionately influencing the overall evaluation. In addition to aggregate performance metrics over this subset, we report results at the individual dataset level. This detailed analysis enables us to examine variability in model performance across different anomaly types, domains, and signal characteristics. 

To maintain comparability with prior studies, we also evaluate our methods on the original UCR anomaly detection dataset without the curation applied in TSB-AD. 

Similar to~\cite{liuElephantRoomReliable2025}, we apply per-series z-normalization as a preprocessing step. We adopt an offline detection setting where the complete time series is available prior to detection; hence, normalization statistics are computed over the entire series. Following the protocol of TSB-AD, models are trained using the benchmark-provided split, but anomaly scores are computed and evaluated for the full series. This setup differs from online anomaly detection, which is not the focus of this work.

\subsection{Models Selection}
We evaluate linear models, deep learning architectures, and PCA-based baselines.

\subsubsection{PCA as a non-learning baseline}
Interestingly, PCA-based approaches can be used in two different ways. The first is the sub-PCA approach, which achieves the best performance~\cite{liuElephantRoomReliable2025} on the TSB-AD benchmark. This method consists of computing the distance between each point and the principal components (see more details in~\cite{shyu2003novel}). The second approach formulates the problem as a basic reconstruction task~\cite{takeishi2019shapley}. Using a truncated SVD, subsequences are projected into a lower-dimensional space and then reconstructed back into the original space. The anomaly score is given by the reconstruction error, which reflects the loss of information caused by the dimensionality reduction.
To evaluate reconstruction-based methods in a way consistent with the deep learning approach, we apply truncated SVD to subsequences from the training set to construct a representative basis. This basis is then used to reconstruct the test set, and the reconstruction error is measured to assess anomalies. Prior to SVD, all subsequences are standardized using the mean and standard deviation of the training set; the same standardization parameters are applied to the test set.

In our experiments, we use both approaches as baselines: the first for its strong empirical performance, and the second as a non-learning-based reconstruction method. To differentiate, we will denote the second approach as the Reconstruction-PCA (\pca).

\subsubsection{DLinear Model as a Learning Baseline}

To assess the actual benefits of more complex deep learning architectures, we use the DLinear model~\cite{zeng2023transformers} as a simple baseline.

In reconstruction-based anomaly detection, linear models could, in theory, converge to a trivial identity mapping. In other words, the model would simply reproduce the input, leading to near-zero reconstruction error and making anomaly detection ineffective. However, this behavior is not observed in practice under standard training conditions, because reaching such an identity mapping would require a large number of training epochs and very specific optimization dynamics.

One of the main strengths of DLinear is its simplicity, resulting in fast training and inference. It also limits the number of hyperparameters that need to be tuned. The only architectural parameter we adjust is the kernel size of the moving average used in the decomposition step.

Overall, DLinear serves as a simple and fast baseline, allowing us to evaluate whether the added complexity of more sophisticated models leads to meaningful improvements.

\subsubsection{Deep Learning Models}

We select a set of deep learning models that represent different reconstruction paradigms and inductive biases for time series modeling. First, we include a classical autoencoder as a fundamental baseline. This model consists of a feedforward encoder-decoder architecture trained to compress and reconstruct input subsequences. Despite its simplicity and the absence of explicit temporal modeling, it provides a useful reference point to assess whether more sophisticated architectures effectively exploit temporal dependencies.

Second, we consider models specifically designed to capture temporal structure. We include TimesNet~\cite{wu2022timesnet}, which is explicitly periodicity-aware and models temporal variations through a two-dimensional representation, enabling the capture of multi-periodic patterns in the time domain. In addition, we evaluate several Transformer-based architectures. First, we use the basic Transformer architecture. We further include advanced Transformer variants such as AutoFormer~\cite{wu2021autoformer} and FEDFormer~\cite{zhou2022fedformer}, which incorporate decomposition-based attention and, in the case of FEDformer, frequency-domain representations to improve the modeling of long-range dependencies.

These models allow us to compare reconstruction-based anomaly detection performance across different inductive biases: compression-based representations (autoencoder), periodicity-aware modeling (TimesNet), frequency-domain modeling (FEDformer), and attention-based temporal modeling (Transformer variants). This selection enables a controlled evaluation of whether increased architectural complexity and specialized temporal modeling lead to consistent improvements over simpler baselines.

\subsection{Hyperparameter Tuning}

We follow the evaluation protocol of the TSB-AD benchmark, which reserves 15\% of the time series for the hyperparameter tuning. Details on the split can be found in~\cite{liuElephantRoomReliable2025}. In contrast to approaches that rely on default configurations, we perform a systematic hyperparameter optimization covering both architectural and training choices. This is motivated by two primary observations: 1) default configurations are often designed for multivariate settings and may be suboptimal for the univariate cases considered in this work; 2) hyperparameter selection significantly impacts performance and should be treated as an integral part of the evaluation protocol.

To maintain a tractable search space, architectural parameters are not tuned independently. Instead, each model is evaluated using three predefined architecture presets: \textit{small}, \textit{medium}, and \textit{large}. These presets group key structural parameters, such as hidden dimensions, number of layers, and attention sizes, enabling a controlled comparison across different model capacities without requiring an exhaustive search over all possible configurations.

In addition to architectural choices, we consider two optimization schemes: a constant learning rate and an adaptive schedule based on \texttt{ReduceLROnPlateau}. The latter adapts the learning rate based on validation performance with a patience of 5 epochs and a reduction factor of 0.5. The choice of tuning with a constant scheduler is motivated by preliminary observations showing that inappropriate scheduling can hinder convergence and significantly degrade performance. Since the loss landscape and reconstruction requirements vary across models, we tune the number of training epochs per model among $\{10, 20, 30, 50\}$, and test initial learning rates of $\{10^{-2}, 10^{-3}, 10^{-4}\}$. Furthermore, we evaluate three subsequence lengths: 32, 64, and 96.

To account for stochastic variability, each configuration is evaluated using three random seeds ($0, 1, 2$), and then the results are averaged across runs. Model selection is performed by maximizing the AUC-PR on the validation split. To avoid over-interpreting marginal differences, AUC-PR values are rounded to two decimal places, and improvements below 1\% are considered negligible. When multiple configurations achieve similar performance, we select the one with the lower computational cost, favoring smaller architecture presets or simpler optimization schemes.

With two inference strategies, two learning rate schedulers, three initial learning rates, three subsequence lengths, and three architecture sizes, the search space results in 108 trials per model. The full hyperparameter search space, including the definition of architecture presets and training configurations for each model, is provided in the supplementary material (Tab.~\ref{tab:model_search_space} and Tab.~\ref{tab:model_specific_presets}). We always use the Adam optimizer with a batch size of 1024.

For \pca{}, we perform the hyperparameter tuning using the same three subsequence lengths, and test with three different explained variance percentages \(\{0.25, 0.5, 0.75\}\) for the lower space dimension. 

\subsection{Model Evaluation}

All models are evaluated using standard reconstruction-based anomaly detection metrics. For each time series, a model assigns an anomaly score at each time step based on reconstruction error. Performance is quantified primarily using the area under the precision-recall curve (AUC-PR), supplemented by the area under the receiver operating characteristic curve (AUC-ROC) and F1-score to provide a more complete picture of detection quality. We use the standard F1 score without any adjustment, such as the point adjustment, which is known to be biased~\cite{liuElephantRoomReliable2025}. 

To account for stochastic variability in model training, each experiment is repeated using five different random seeds ($3,4,5,6,7$) different from the one used for the hyperparameter tuning. Results are averaged across these five runs.  Note that we do not to include range-based metrics such as volume under the surface~\cite{boniolVUSEffectiveEfficient2025}, and this choice is motivated by their high execution time, which significantly increases the cost of the evaluation, especially in our context, where it is carried out on five seeds.

\section{Results}
\label{sec:results}

\subsection{Setup}
The experiments were implemented in Python, with all deep learning components developed using PyTorch. The deep learning models were adopted from the Time Series Library\footnote{\url{https://github.com/thuml/Time-Series-Library}}~\cite{wangDeepTimeSeries2024}. For the evaluation of model performance, including the computation of relevant metrics, we used the TSB-AD package\footnote{\url{https://github.com/thedatumorg/TSB-AD}}~\cite{liuElephantRoomReliable2025}. All computations were made on a NVIDIA V100 GPU. The code of all experiments is available on GitHub\footnote{\url{https://github.com/GuillaumeCld/DeepTSAD_eval}}.

\subsection {Impact of Windowing Strategy}

We begin by quantifying how the windowing strategy in the inference phase affects reconstruction-based anomaly detection. Table~\ref{tab:inference_delta} reports benchmark-level averages on the 350 evaluation time series of TSB-AD under two windowing strategies: disjoint ($s=w$) and overlapping ($s=1$). Across all models and all three metrics (AUC-PR, AUC-ROC, and Standard-F1), overlapping inference yields higher scores, with AUC-PR absolute gains ranging from about +0.02 to +0.10 on average, and relative (percentage) gains up to 28\%.

\begin{table*}[hbt]
\centering
\caption{Average performance under disjoint and overlapping inference on the 350 evaluation time series of TSB-AD. $\Delta\%$ denotes the relative change in percentage. Results are ordered by AUC-PR gains.}
\label{tab:inference_delta}
\small
\setlength{\tabcolsep}{6pt}
\begin{tabular}{l|ccc|ccc|ccc}
\toprule
\multirow{2}{*}{Model} 
& \multicolumn{3}{c|}{AUC-PR} 
& \multicolumn{3}{c|}{AUC-ROC} 
& \multicolumn{3}{c}{Standard-F1} \\
\cmidrule(lr){2-4}\cmidrule(lr){5-7}\cmidrule(lr){8-10}
& Disjoint & Overlap & $\Delta\%$ 
& Disjoint & Overlap & $\Delta\%$ 
& Disjoint & Overlap & $\Delta\%$ \\
\midrule
TimesNet     & 0.32& 0.41& +28.1\%& 0.72& 0.79&  +9.7\%& 0.36& 0.45& +25.0\%\\
DLinear      & 0.33& 0.42& +27.3\%& 0.73& 0.80&  +9.6\%& 0.37& 0.46& +24.3\%\\
AutoEncoder  & 0.37& 0.47& +27.0\%& 0.78& 0.86& +10.3\%& 0.41& 0.51& +24.4\%\\
Rec-PCA      & 0.31& 0.37& +19.4\%& 0.71& 0.76&  +7.0\%& 0.36& 0.41& +13.9\%\\
FEDFormer    & 0.34& 0.37&  +8.8\%& 0.69& 0.71&  +2.9\%& 0.38& 0.41&  +7.9\%\\
AutoFormer   & 0.33& 0.35&  +6.1\%& 0.69& 0.69&  +0.0\%& 0.37& 0.39&  +5.4\%\\
Transformer  & 0.36& 0.38&  +5.6\%& 0.74& 0.76&  +2.7\%& 0.40& 0.42&  +5.0\%\\
\bottomrule
\end{tabular}
\end{table*}

The magnitude of this improvement, however, depends on the model family. Lightweight baselines and classical methods benefit the most in relative terms: TimesNet and DLinear improve AUC-PR by about +28\% and +27\%, respectively, while Rec-PCA gains roughly +19\%. The AutoEncoder also shows a strong effect, with +27\% AUC-PR and +24\% Standard-F1, suggesting that disjoint segmentation can substantially deteriorate reconstruction errors. By contrast, Transformer-based variants (Transformer, FEDFormer, AutoFormer) exhibit more modest relative gains in AUC-PR (about +6\% to +9\%), and remain less competitive in absolute performance even under overlapping inference. Interestingly, with overlap enabled, Rec-PCA reaches a similar AUC-PR level (0.37) to FEDFormer (0.37) and comes close to Transformer (0.38), making it broadly comparable to these Transformer variants on this metric.

To complement aggregated averages, Table~\ref{tab:overlap_delta_stats} evaluates the strategy effect per time series. For each model and series, we compute the paired improvement
$\Delta=\mathrm{AUC\mbox{-}PR}_{\text{overlap}}-\mathrm{AUC\mbox{-}PR}_{\text{disjoint}}$ and summarize $\Delta$ by its mean, median, and the fraction of series with $\Delta>0$. Overlapping inference improves performance on a large majority of series for every model (e.g., 88.9\% for AutoEncoder and 81.1\% for TimesNet), and the improvements remain statistically significant under a paired Wilcoxon signed-rank test with Holm correction ($\alpha=0.05$)~\cite{wilcoxon1992individual,holm1979simple}. The \(p\)-value of the tests are reported in the table.

\begin{table}[hbt]
\centering
\caption{Per-series absolute AUC-PR gain \(\Delta\) from overlapping inference. Results are ordered by gains.}
\label{tab:overlap_delta_stats}
\small
\begin{tabular}{lcccc}
\toprule
Model & Mean $\Delta$ & Median $\Delta$ & \% Improved & $p$-value  \\
\midrule
AutoEncoder & 0.10 & 0.06 & 88.9 & $<1e^{-4}$  \\
TimesNet    & 0.09 & 0.06 & 81.1 & $<1e^{-4}$  \\
DLinear     & 0.09 & 0.03 & 71.1 & $<1e^{-4}$  \\
Rec-PCA     & 0.06 & 0.02 & 80.6 & $<1e^{-4}$  \\
FEDFormer   & 0.03 & 0.00 & 69.1 & $<1e^{-4}$  \\
AutoFormer  & 0.02 & 0.00 & 67.1 & $<1e^{-4}$  \\
Transformer & 0.02 & 0.00 & 56.6 & $\approx 1e^{-4}$  \\
\bottomrule
\end{tabular}
\end{table}

\begin{figure}[hbt]
     \centering
     \begin{subfigure}[t]{\linewidth}
         \centering
         \includegraphics[width=\linewidth]{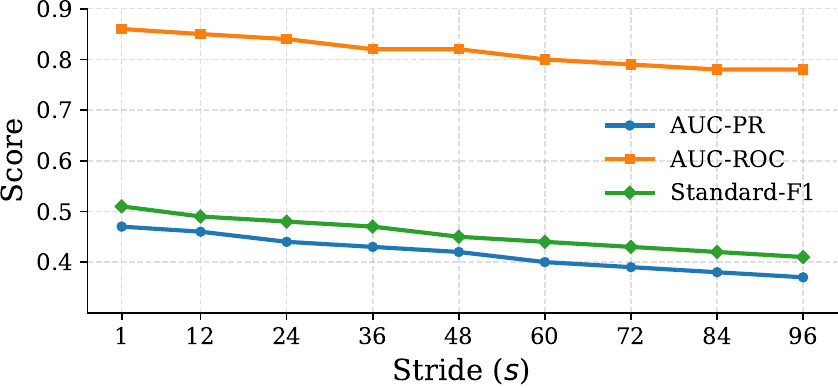}
         \caption{Average AUC-PR, AUC-ROC, F1-score}
         \label{fig:stride_metric}
     \end{subfigure}
     \hfill
     \begin{subfigure}[t]{\linewidth}
         \centering
         \includegraphics[width=\linewidth]{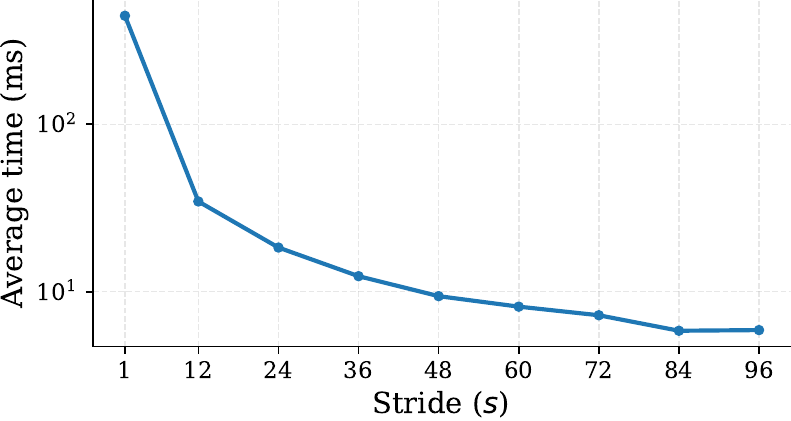}
         \caption{Average inference execution time (in ms)}
         \label{fig:stride_time}
     \end{subfigure}
    \caption{Average performance/runtime tradeoff as a function of the stride \(s\) in inference for the best AutoEncoder configuration on the evaluation dataset.}
    \label{fig:slide_stride}
\end{figure}

Finally, we examine the performance/runtime trade-off by sweeping the inference stride for the best AutoEncoder configuration, therefore isolating the impact of the stride. Figure~\ref{fig:slide_stride} shows that detection quality monotonically degrades as the stride increases from $s=1$ (maximal overlap) to $s=w$ (disjoint windows), while execution time decreases accordingly. Notably, the AUC-PR and F1-score decrease by up to 10\%, while the AUC-ROC decreases by up to 8\%, consistent with the results reported in Table~\ref{tab:inference_delta}.

Overall, these results demonstrate that overlapping inference provides an important and non-trivial performance boost across heterogeneous reconstruction models. In several cases, the gain from overlapping inference is comparable to (or larger than) the gap between model families under disjoint inference, highlighting that inference settings can meaningfully influence comparative conclusions.

Since the overlapping window strategy consistently yields the best results, all subsequent analyses will be carried out using overlapping windows with \(s=1\).

\subsection{Model Performance Comparison}

We now turn to the comparison of model performance across the selected architectures. Table~\ref{tab:tsb_uad_metrics} summarizes the anomaly detection results obtained for each model using multiple evaluation metrics. To improve the robustness of the evaluation, every experiment was repeated five times with different random seeds. For each time series, we first computed the mean and standard deviation of the metrics across the five runs. The reported benchmark-level results correspond to the average of these per-series means, together with the average of the associated standard deviations, thereby reflecting the typical performance and stability of each model on the benchmark.

Among all evaluated methods, the AutoEncoder achieves the strongest overall performance, obtaining the highest scores on all metrics with an AUC-PR of 0.47, an AUC-ROC of 0.86, and an F1-score of 0.51. Notably, it clearly outperforms both the original Sub-PCA baseline and the \pca{} implementation used in our experiments. DLinear also delivers competitive results, ranking second across all metrics despite its comparatively simple architecture. In contrast, more sophisticated transformer-based approaches such as FEDFormer and AutoFormer obtain weaker performances overall, suggesting that higher architectural complexity does not necessarily translate into better anomaly detection capability in the univariate setting. Most of the evaluated reconstruction pipelines achieve average performance comparable to or better than Sub-PCA, which was the top-performing method in the original benchmark. This indicates that the reconstruction-based paradigm can be regarded as one of the most effective approaches for anomaly detection within the context of this benchmark.

\begin{table}[htb]
    \centering
    \caption{Average anomaly detection performance. Each experiment is repeated five times; results report the average of per-series means and standard deviations across the five seeds in brackets. }
    \begin{tabular}{c|c c c}
    \toprule
        Model       &  AUC-PR & AUC-ROC &  F1 \\\midrule
        Sub-PCA     & \res{0.37}{0.00}    & \res{0.71}{0.00}    & \res{0.42}{0.00}\\
        \pca{}      & \res{0.37}{0.00} & \res{0.76}{0.00} & \res{0.41}{0.00} \\
        DLinear     &   \res{0.42}{0.01}&  \res{0.80}{0.00}&  \res{0.46}{0.01}         \\
        AutoEncoder &  \bf \res{0.47}{0.02}& \bf\res{0.86}{0.01} & \bf\res{0.51}{0.02} \\
        TimesNet    &  \res{0.40}{0.04}& \res{0.79}{0.03} & \res{0.44}{0.04}            \\
        Transformer &  \res{0.38}{0.04}& \res{0.76}{0.04}& \res{0.42}{0.04}            \\
        FEDFormer    &  \res{0.37}{0.02}& \res{0.71}{0.02}& \res{0.41}{0.02}            \\
        AutoFormer  &  \res{0.35}{0.02}& \res{0.69}{0.03}& \res{0.39}{0.02}            \\
    \bottomrule
    \end{tabular}
    \label{tab:tsb_uad_metrics}
\end{table}

\begin{figure}[htb]
    \centering
    \includegraphics[width=\linewidth]{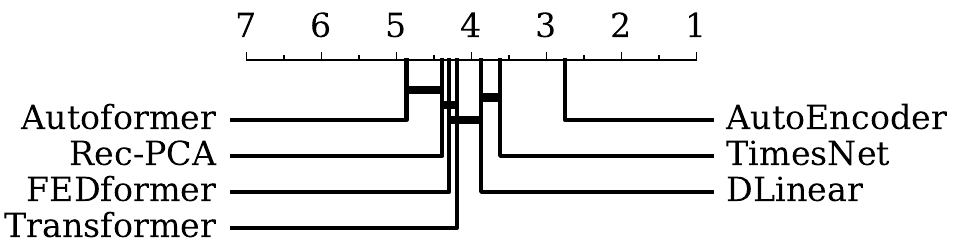}
    \caption{Critical difference diagram computed with the Friedman test followed by a post-hoc Wilcoxon test (with \(\alpha = 0.05\)) on the averaged AUC-PR of each of the 350 evaluation time series}
    \label{fig:critical_diag}
\end{figure}

We are now looking at the significance of the difference. Figure~\ref{fig:critical_diag} displays a critical difference diagram computed with the Friedman test followed by a post-hoc Wilcoxon test~\cite{wilcoxon1992individual} as in~\cite{boniolVUSEffectiveEfficient2025}. AutoEncoder achieved the best average rank, while AutoFormer obtained the worst. However, the post-hoc Wilcoxon analysis indicates that many pairwise differences are not statistically significant, as shown by the connected groups in the critical difference diagram. AutoEncoder is the top-ranked method and is significantly better than most alternatives

Overall, these results indicate that anomaly detection performance is not solely a function of model complexity. Efficient, lightweight architectures, such as DLinear, frequently match or even surpass more sophisticated models, underscoring the critical importance of establishing strong baselines and 
adhering to rigorous evaluation protocols. Furthermore, high-capacity models, specifically Transformers and TimesNet, exhibit significant performance variability, with an average standard deviation of 3 to 4\% per series. This lack of robustness suggests difficulties in optimizing complex models using a static set of hyperparameters across a highly diverse range of time series. These findings emphasize: (i) the necessity of reporting results across multiple seeds to ensure fair comparative analysis, and (ii) the need for more robust approaches that guarantee reliability beyond controlled benchmarks. Crucially, these results highlight that simple reconstruction-based methods remain highly effective on this benchmark.

When comparing our findings with the public TSB-AD ranking\footnote{Accessible online on \url{https://thedatumorg.github.io/TSB-AD/} (consulted on 31/03/2026)} reported in~\cite{liuElephantRoomReliable2025}, we observe substantial performance gains obtained through careful hyperparameter tuning and the selection of appropriate inference strategies. For example, TimesNet improves from an AUC-PR of 0.18 to 0.40, an AUC-ROC of 0.61 to 0.79, and an F1-score of 0.24 to 0.44.

These observations underline the importance of conducting controlled and systematic evaluations. By jointly exploring hyperparameters and inference procedures, our study provides a more comprehensive assessment of model capabilities and offers additional insights into the effectiveness of deep learning approaches for univariate anomaly detection.

However, while the benchmark-level averages provide a useful global overview, they can also conceal substantial variations in performance across individual datasets. In practice, some models exhibit markedly different behaviors depending on the characteristics of the underlying time series, such as anomaly type, temporal dynamics, or noise patterns. To better illustrate these discrepancies and move beyond aggregated results, we next examine model performance at the dataset level.

\subsection{Dataset-Level Analysis}

Model performance varies considerably across datasets, indicating that benchmark-level averages alone may mask important disparities. Some datasets consistently yield high detection scores for most methods, while others remain challenging regardless of the architecture considered. This highlights the importance of conducting analyses at the dataset level in addition to reporting aggregated metrics.

\begin{figure}[htb]
    \centering
    \includegraphics[width=\linewidth]{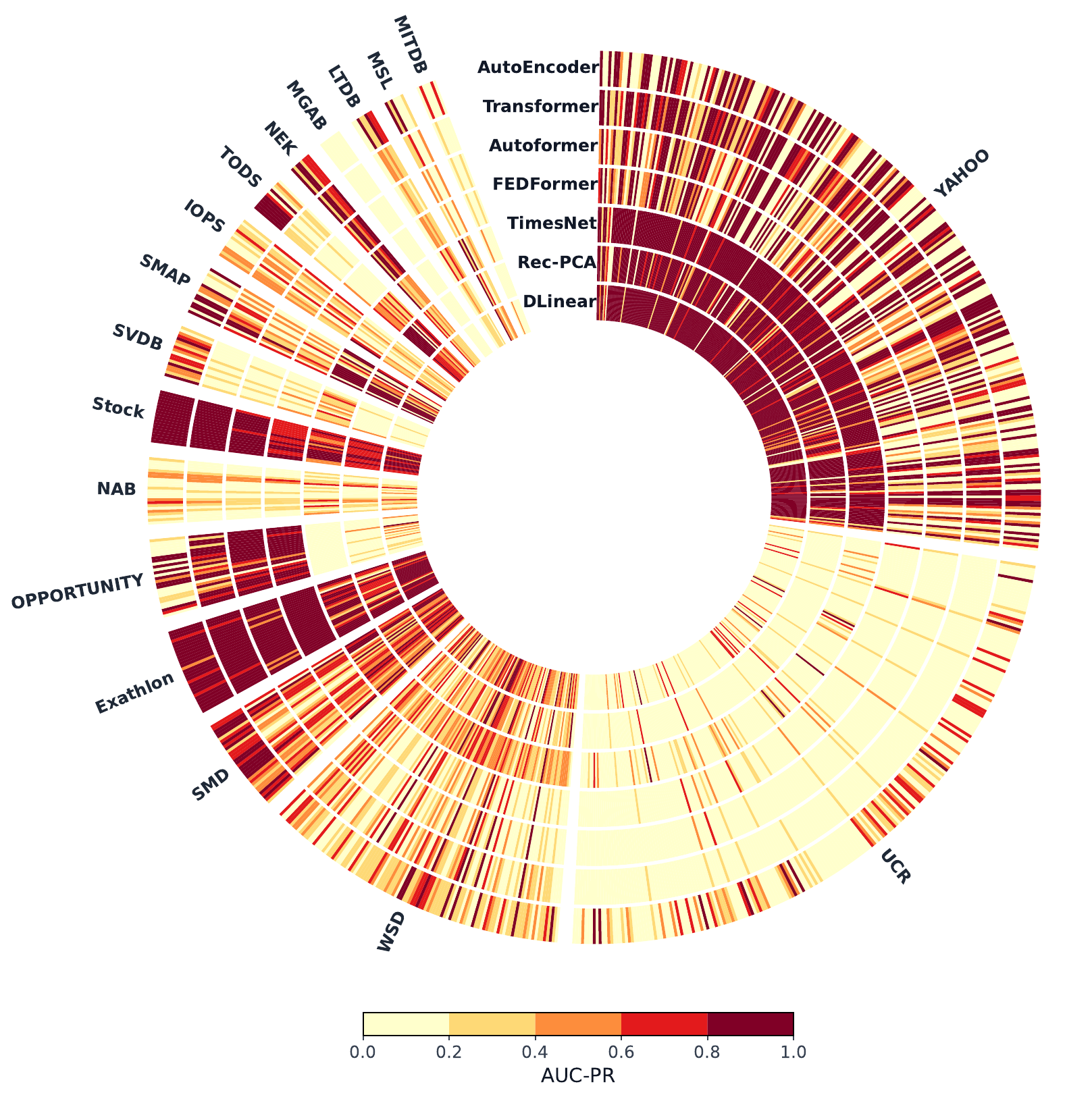}
    \caption{Polar heatmap comparing anomaly detection performance (AUC-PR) across datasets and models. Each concentric ring corresponds to a model (inner to outer: DLinear, \pca{}, TimesNet, FEDFormer, AutoFormer, Transformer, AutoEncoder), and each angular segment represents a time series instance grouped by dataset; warmer colors indicate higher AUC-PR.}
    \label{fig:model_comparison_polar_heatmap}
\end{figure}

Figure~\ref{fig:model_comparison_polar_heatmap} presents a fine-grained visualization of anomaly detection performance across datasets and models using a polar heatmap representation. In contrast to aggregated statistics, this visualization exposes performance distributions at the individual time-series level, enabling a direct assessment of robustness, consistency, and variability across heterogeneous anomaly detection scenarios. Several important observations can be drawn from this figure.

First, the heatmap highlights a strong dataset dependency. Some datasets, such as Stock, Exathlon, and partially YAHOO, exhibit consistently high AUC-PR values across most models, visible through large contiguous dark-red regions. In contrast, datasets such as UCR, MGAB, and MITDB contain predominantly low-intensity regions, indicating that these datasets remain challenging regardless of the selected architecture.

Then, the figure emphasizes the substantial variability observed even within the same dataset. For instance, on the YAHOO dataset, the AutoEncoder model display many high values (above 0.8) and many low values (below 0.2), resulting in an average performance, while very few average values are observable. Providing more insights into the model performance.

Finally, the heatmap also illustrates that simpler approaches remain highly competitive in several scenarios. \pca{} achieves strong performance on datasets such as Stock and TODS, demonstrating that increased architectural complexity does not systematically translate into better anomaly detection capability. Overall, these observations support the importance of comprehensive multi-dataset evaluations and carefully tuned baselines when assessing anomaly detection methods.

To support the analysis with values, Table~\ref{tab:per_dataset_models_full} reports the average AUC-PR obtained on each dataset by all methods. Overall, the AutoEncoder achieves the best average performance (0.48), outperforming DLinear (0.42), TimesNet (0.39), and \pca{} (0.36). However, despite this superior average score, the results do not reveal a universally dominant model. Indeed, DLinear achieves the best performance on several large and practically relevant datasets, including WSD and SMD. TimesNet performs particularly well on the YAHOO dataset, but such as DLinear. Conversely, \pca{} remains surprisingly competitive on datasets such as TODS. As visible in Figure~\ref{fig:model_comparison_polar_heatmap}, transformer-based architecture perform significantly better on the OPPORTUNITY dataset with score of 0.84 for both AutoFormer and FEDFormer.

\begin{table*}[htb]
    \centering
    \begin{tabular}{l|c|ccccccccc}
    \toprule
    Dataset & Cardinal & \pca & DLinear & AE & TimesNet & Transformer & AutoFormer & FEDFormer \\
    \midrule
    YAHOO       & 254 & 0.81 & \bf0.91 & 0.47 & \bf0.91 & 0.61 & 0.47 & 0.49\\
    UCR         & 222 & 0.04 & 0.10 &\bf 0.21 & 0.13 & 0.03 & 0.04 & 0.05 \\
    WSD         & 106 & 0.37 & \bf0.47 & 0.36 & 0.43 & 0.32 & 0.24 & 0.24 \\
    SMD         & 33 & 0.61  & 0.69 & \bf0.70 & 0.62 & 0.48 & 0.40 & 0.42\\
    Exathlon    & 30 & 0.70 & 0.89 & 0.90 & 0.79 & 0.90 & 0.85 & \bf0.93\\
    OPPORTUNITY & 27 & 0.10 & 0.20 & 0.40 & 0.05 & 0.69 & \bf0.84 & \bf0.84\\
    NAB         & 23 & 0.24 & 0.27 & 0.27 & \bf0.28& 0.23 & 0.20 & 0.22\\
    Stock       & 18 & 0.77 & 0.82 & \bf0.99 & 0.73& 0.93 & 0.87 & 0.80\\
    SVDB        & 18 & 0.10 & 0.11 & \bf0.51 & 0.32& 0.16 & 0.14 & 0.20\\
    SMAP        & 17 & 0.63 & 0.58 & 0.48 & \bf0.66& 0.43 & 0.31 & 0.33\\
    IOPS        & 15 & 0.26 & 0.35 & 0.30 & 0.31& \bf0.37 & 0.33 & 0.34\\
    TODS        & 13 & \bf0.74 & 0.62 & 0.62 & 0.48& 0.20 & 0.13 & 0.14\\
    NEK         & 8 & 0.47 & 0.38 & 0.71 & 0.49& 0.70 & 0.72 & \bf0.76\\
    MGAB        & 8 & 0.01 & 0.00 & \bf0.14 & 0.03& 0.00 & 0.01 & 0.00\\
    LTDB        & 8 & 0.24 & 0.21 & \bf0.51 & 0.40& 0.29 & 0.27 & 0.32\\
    MSL         & 7 & 0.41 & 0.43 & 0.38 & \bf0.48& 0.28 & 0.28 & 0.26\\
    MITDB       & 7 & 0.06 & 0.12 & \bf0.27 & 0.14& 0.12 & 0.12 & 0.15\\
    \midrule
    \midrule
    Average               & - & 0.36 & 0.42 & \bf0.48 & 0.39 & 0.40 & 0.36 & 0.38\\
    Standard deviation    & - & 0.29 & 0.30 & 0.26 & 0.25 & 0.30 & 0.31 & 0.30\\
    \bottomrule
    \end{tabular}
    \caption{Average AUC-PR per model per dataset of TSB-AD.}
    \label{tab:per_dataset_models_full}
\end{table*}

Another important observation is the substantial variability across datasets. Performance ranges from almost random detection on datasets such as MGAB to excellent detection on datasets such as Stock and Exathlon. This wide dispersion illustrates the intrinsic heterogeneity of anomaly detection benchmarks and further supports the necessity of broad multi-dataset evaluation protocols when assessing model effectiveness.

\subsection{Zoom on the UCR Dataset}
To provide more insights on reconstruction based anomaly detection, we now focus on the UCR dataset, an important benchmark given the number and diversity of time series. Each of the 250 time series is accompanied by detailed provenance information, which increases confidence in the reliability of the ground truth annotations. The datasets span a wide variety of domains, including cardiology, industrial systems, medicine, zoology, meteorology, and human behavior.

To ensure a fair comparison between models and paradigms on this archive, we follow the evaluation protocol introduced by its creators and evaluate all 250 time series in their original form. In this setting, each dataset contains exactly one anomaly, represented by an anomalous interval \( [t_s, t_e] \), and the objective of an algorithm is to predict its location.

Let
\[
L = t_e - t_s
\]
denote the anomaly length. Following the original protocol, a prediction \( \hat{t} \) is considered correct if it falls within a tolerance region extending by \( L \) samples before and after the anomalous interval, i.e.,
\[
\hat{t} \in [t_s - L,\; t_e + L].
\]
When \( L < 100 \), the tolerance is set to 100 to avoid overly strict evaluation on short anomalies. This defines a relaxed localization criterion that rewards coarse anomaly localization.

However, this evaluation protocol is not fully adapted to reconstruction-based anomaly detection methods operating on sliding subsequences, especially when the considered subsequence lengths  are less than \(L\) (at most 96 in this study). Thus, for a fair evaluation, the tolerance window length should not be greater than the considered subsequence length.

To better reflect this setting, we introduce two additional localization criteria tailored to subsequence-based detection:

\begin{itemize}
    \item \textbf{Semi-strict score:} a prediction is considered correct if it falls within a tolerance region extending by \( w \) samples before and after the anomalous interval, i.e.,
    \[
    \hat{t} \in [t_s - w,\; t_e + w].
    \]
    This criterion accounts for detected points belonging to any reconstructed subsequences overlapping with the anomaly. 

    \item \textbf{Strict score:} a prediction is considered correct only if it lies inside the annotated anomalous interval itself, i.e.,
    \[
    \hat{t} \in [t_s,\; t_e].
    \]
    This criterion measures the ability of the method to precisely localize the anomaly without any additional tolerance outside the ground-truth anomalous region.
\end{itemize}

Together, these three metrics provide complementary perspectives on localization performance: the original score serves as a comparable baseline but might inflate the results, the semi-strict score evaluates localization at the subsequence resolution, and the strict score measures precise alignment with the annotated anomalous interval.

For the sake of consistency, we keep the hyperparameters previously selected. It is to be noted that, among the 48 times series used for the hyperparameter tuning, six come from the UCR dataset. Therefore, the results are conservative relative to UCR-specific tuning reducing the risk of benchmark-specific overfitting.

The results reported in Table~\ref{tab:ucr_strict} show that performance under this metric is generally higher than what is suggested by the AUC-PR values presented in Figure~\ref{fig:model_comparison_polar_heatmap} and Table~\ref{tab:per_dataset_models_full}. This difference stems from the nature of the evaluation protocol, which focuses on localizing a single anomaly rather than assessing the full ranking of anomaly scores over time. However, despite this apparent improvement, the results remain far from perfect, indicating that the task is still challenging. For comparison, Table~\ref{tab:ucr_baseline} reports the performance of additional methods from the literature, as presented in recent works~\cite{imamuraGeneralizedDiscordsTime2025a,tafazoliC22MPMarriageCatch222024}.

\begin{table}[htb]
    \centering
    \caption{Performance comparison on the full UCR archive using the strict and semi-strict score for the reconstruction models.}
    \begin{tabular}{c|ccc}
    \toprule
    Method & UCR score& Semi-strict & Strict\\
    \midrule
    AutoEncoder & \rpm{53.1}{0.3} & \rpm{53.1}{0.3} & \rpm{43.0}{0.2} \\
    TimesNet    & \rpm{52.3}{1.5} & \rpm{52.1}{1.7} & \rpm{42.1}{1.4} \\
    DLinear     & \rpm{38.3}{0.5} & \rpm{38.3}{0.5} & \rpm{25.0}{0.4} \\
    \pca{}      & \rpm{25.2}{0.0} & \rpm{25.2}{0.0} & \rpm{18.4}{0.0} \\
    Transformer & \rpm{20.7}{1.2} & \rpm{19.9}{0.9} & \rpm{17.2}{0.7} \\
    FEDFormer   & \rpm{18.2}{0.9} & \rpm{17.4}{0.5} & \rpm{13.8}{1.0} \\
    AutoFormer  & \rpm{17.2}{0.8} & \rpm{16.3}{1.0} & \rpm{13.9}{0.3} \\
    \bottomrule
    \end{tabular}
    \label{tab:ucr_strict}
\end{table}

Overall, AutoEncoder and TimesNet achieve the strongest performance among all reconstruction approaches, with UCR scores of 53.1 and 52.3, respectively. These two models maintain strong results under both semi-strict and strict evaluations, with only a moderate degradation under the strict setting (e.g., AutoEncoder drops from 53.1 to 43.0).

DLinear follows as a competitive linear baseline, achieving 38.3 under the UCR score and 25.0 under the strict metric. In contrast, Transformer-based architectures (Transformer, FEDFormer, and AutoFormer) perform substantially worse across all metrics, with UCR scores ranging from 17.2 to $20.7$.

A notable observation is the consistent gap between the UCR and strict scores across all methods. This gap reflects the inherent difficulty of precisely identifying anomaly boundaries under stricter evaluation constraints, which penalize temporal misalignment more heavily. 

GDFlex achieves the highest overall UCR score (72.8), significantly outperforming all reconstruction-based approaches. Similarly, Matrix Profile-based approaches, C$^{22}$MP and DAMP also demonstrate strong performance, with scores above 55, but comparable with the bests reconstruction models. Showing that the reconstruction paradigm can be competitive with the distance paradigm, while not being the state-of-the-art

In contrast, traditional deep learning baselines such as USAD, LSTM-VAE, and TranAD achieve substantially lower scores, with values between 19.0 and 27.6. Our results show that deep learning results can be improved on this dataset, revisiting existing conclusion on the potential of deep learning.

Finally, we can see that the semi-strict score can slightly deteriorate the performance of some models, up to -0.9 for AutoFormer and -0.8 for FEDFormer, illustrating the benefits of introducing a stricter score to avoid inflated performances and perform a fair comparison of approaches that might have different subsequence lengths.

\begin{table}[htb]
    \centering
    \caption{Baselines performance on the full UCR archive using the UCR score, results taken from~\cite{imamuraGeneralizedDiscordsTime2025a} and~\cite{tafazoliC22MPMarriageCatch222024}}
    \begin{tabular}{r|c}
    \toprule
      Method & UCR Score \\
    \midrule
     GDFlex~\cite{imamuraGeneralizedDiscordsTime2025a} & 72.8 \\
     C$^{22}$MP~\cite{tafazoliC22MPMarriageCatch222024} & 56.8 \\
     DAMP~\cite{luDAMPAccurateTime2023}   & 55.6  \\
     NORMA~\cite{boniolUnsupervisedScalableSubsequence2021}  & 47.4  \\
     SCRIMP~\cite{zhuMatrixProfileXI2018} & 41.6  \\
     USAD~\cite{audibertUSADUnSupervisedAnomaly2020a}   & 27.6  \\
     LSTM-VAE~\cite{parkMultimodalAnomalyDetector2018} & 19.8 \\
     TranAD~\cite{tuliTranADDeepTransformer2022b} & 19.0 \\
    \bottomrule
    \end{tabular}
    \label{tab:ucr_baseline}
\end{table}

\section{Complementary Analysis}
\label{sec:complementary}

\subsection{Execution time}

\begin{figure}[htb]
    \centering
    \includegraphics[width=\linewidth]{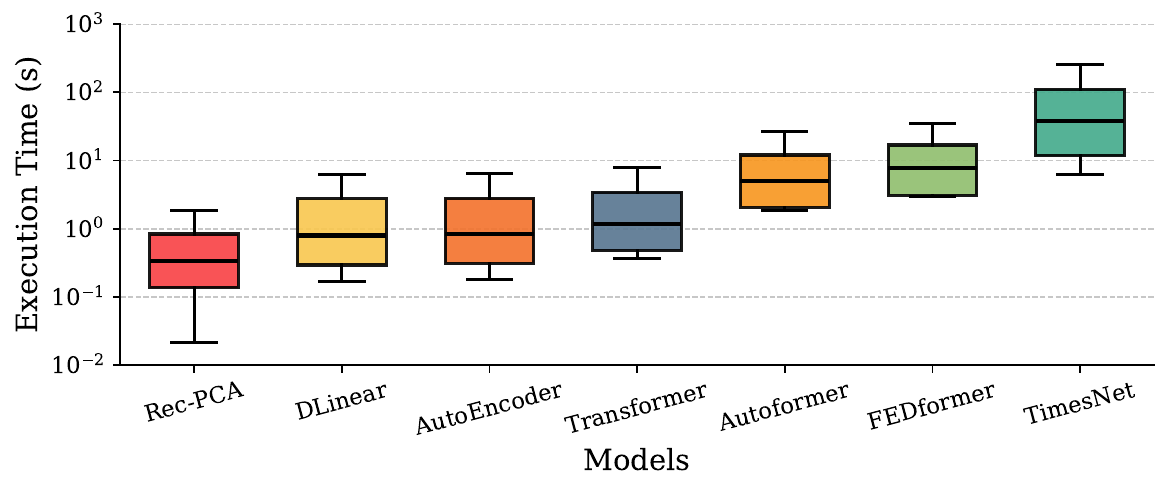}
    \caption{Distribution of the execution time (training+inference) of the reconstruction models on the TSB-AD benchmark.}
    \label{fig:exect_time}
\end{figure}

We also present in Figure~\ref{fig:exect_time} the distribution of execution times for all methods evaluated on the TSB-AD benchmark. The reported execution time includes both training and inference phases. As expected, execution time generally reflects model complexity: \pca{} exhibits the runtime with a median execution time below one second, whereas TimesNet is slower by approximately two orders of magnitude in median runtime. The top-performing methods, notably AutoEncoder and DLinear, achieve median computation times of around one second, demonstrating that these lightweight approaches are computationally efficient despite requiring a training phase.

\subsection{Robustness Evaluation}

We now evaluate sensitivity to hyperparameters and initialization, revealing substantial variability.

\subsubsection{Seed sensitivity}

\begin{figure}[ht]
     \centering
     \begin{subfigure}[t]{\linewidth}
         \centering
         \includegraphics[width=\linewidth]{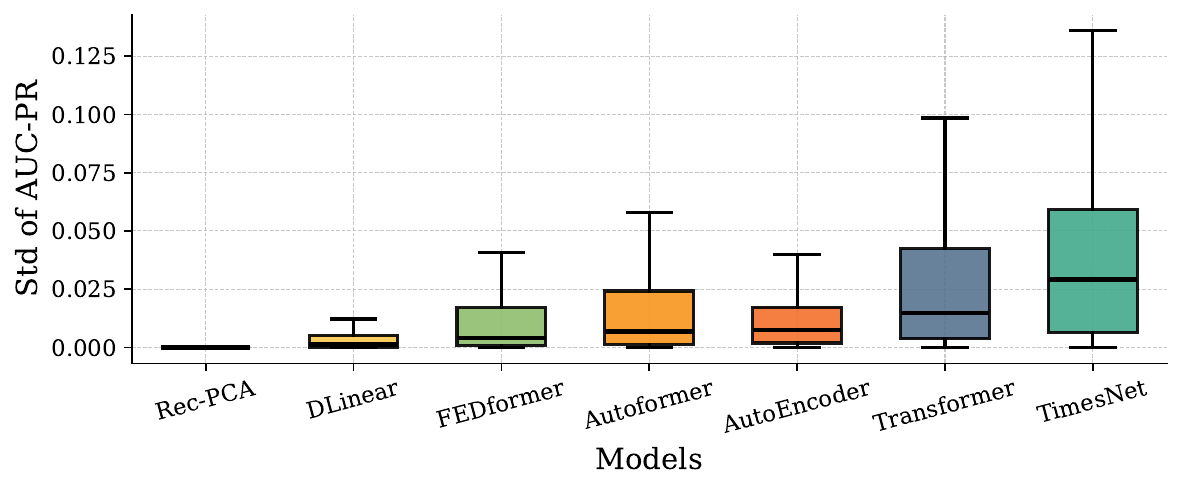}
         \caption{Distribution of standard deviation across seeds}
         \label{fig:robust_seed_std}
     \end{subfigure}
     \hfill
     \begin{subfigure}[t]{\linewidth}
         \centering
         \includegraphics[width=\linewidth]{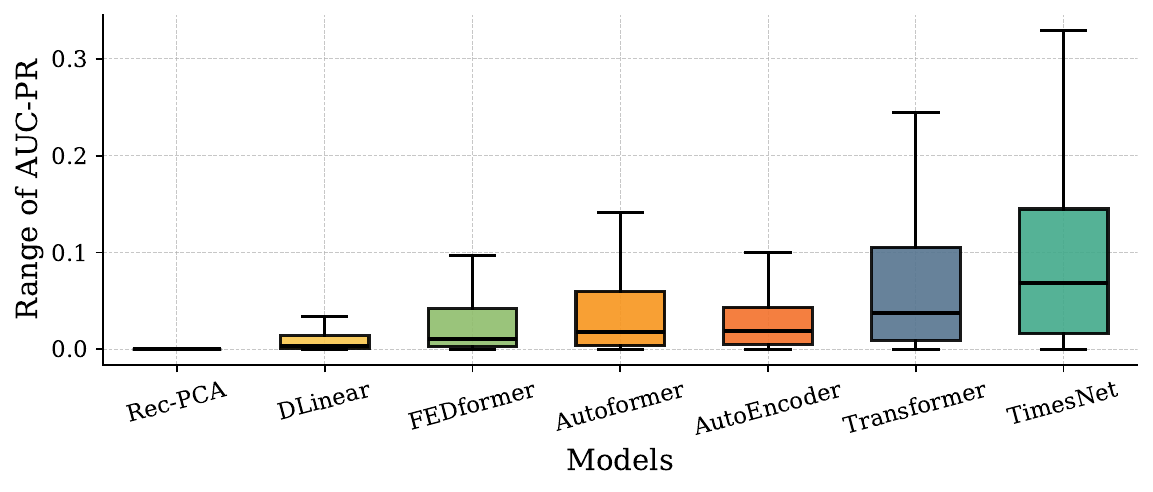}
         \caption{Distribution of range across seeds}
         \label{fig:robust_seed_range}
     \end{subfigure}
    \caption{Evaluation of the robustness to the seed of the reconstruction models on the TSB-AD evaluation benchmark.}
    \label{fig:robust_seed}
\end{figure}

We first evaluate the sensitivity to the seed on the evaluation test of the TSB-AD benchmark across the five evaluation seeds. Figure~\ref{fig:robust_seed} displays for all models the distribution of standard deviation (Fig.~\ref{fig:robust_seed_std}) and range (Fig.~\ref{fig:robust_seed_range}) of the AUC-PR per time series.
Aside from \pca{}, which is deterministic, DLinear achieves the smallest standard deviation range, indicating highly stable performance. In contrast, more complex architectures show progressively larger fluctuations. TimesNet consistently yields the highest standard deviations and range, followed by Transformer and AutoEncoder. In the worst case, TimesNet can have a difference of up to 30\% in AUC-PR on two different seeds for the same time series, raising concerns about the architecture's robustness. Surprisingly, FEDFormer and AutoFormer show a noticeably better robustness then the vanilla Transformer.

These observations suggest that simpler or more constrained architectures tend to exhibit more consistent AUC-PR performance across seeds, whereas higher-capacity sequence models may be more sensitive to optimization dynamics and initialization effects. These results highlight the importance of reporting seed-wise robustness when evaluating time-series anomaly detection models to account for model robustness. 

\subsubsection{Hyperparameter sensitivity}
\begin{figure}[htb]
    \centering
    \includegraphics[width=\linewidth]{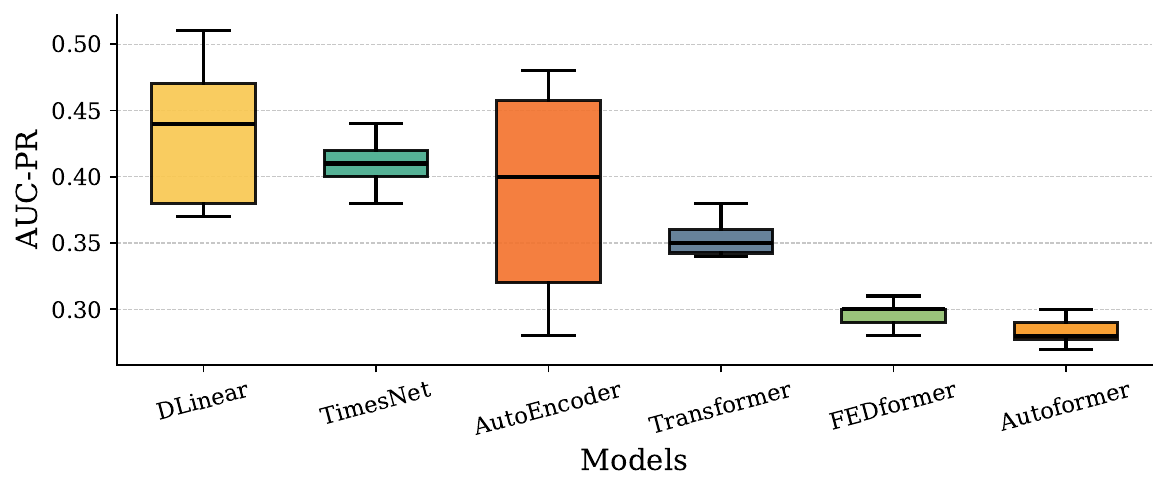}
    \caption{Distribution of AUC-PR across the different hyperparameters set (overlapping only) on the tuning dataset from TSB-AD for all learning models.}
    \label{fig:robust_tuning}
\end{figure}

We now investigate the impact of the hyperparameters selection for the learning models. To simplify the analysis, we focus only on the overlapping inference strategy. Figure~\ref{fig:robust_tuning} displays the distribution of AUC-PR on the hyperparameter tuning dataset on the 54 hyperparmeters sets with overlapping.
Simpler methods (DLinear and AutoEncoder) show a broader variation in performance. While these methods perform well overall, there are parameter settings that perform better than others. Particularly, the AutoEncoder shows a range of 20\% between the best and the worst performing configuration, illustrating the sensitivity of the model to the set of hyperparameters and the need for tuning. Meanwhile, transformer-based models have a range of approximately 5\%, highlighting a better robustness on the tuning dataset; however performing worse than the median of the other models.

\begin{figure}[htb]
    \centering
    \includegraphics[width=\linewidth]{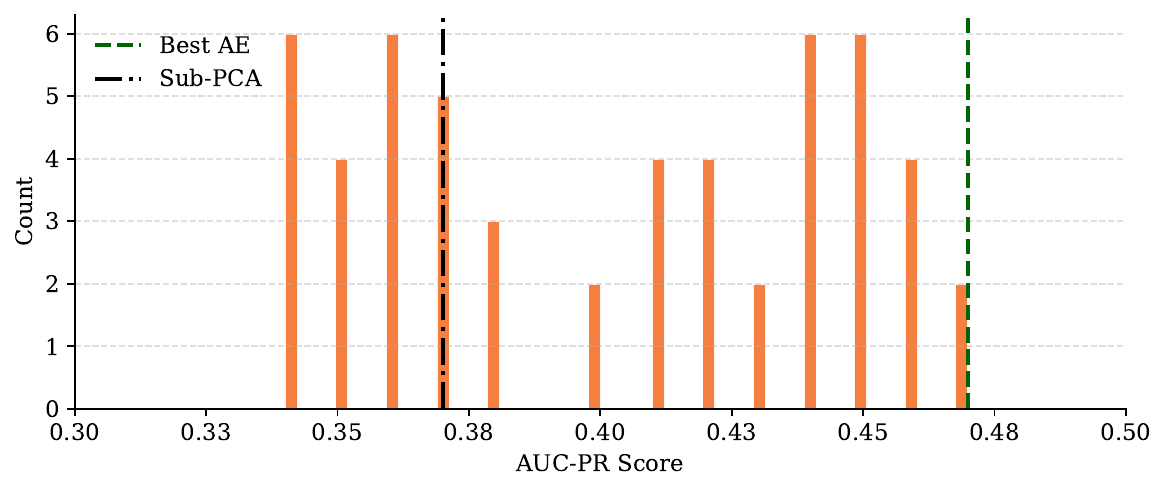}
    \caption{Distribution of AUC-PR across the 54 different combination of hyperparameters (overlapping only) set on the evaluation dataset from TSB-AD for the AutoEncoder.}
    \label{fig:robust_eval_ae}
\end{figure}

In Figure~\ref{fig:robust_eval_ae}, we present the distribution of AUC-PR scores on the TSB-AD evaluation dataset. Due to computational constraints, we restrict the analysis to the AutoEncoder, as it achieved the best overall performance. The green vertical line denotes the best-performing AutoEncoder configuration, corresponding to the model discussed in the previous sections, while the black vertical line represents the Sub-PCA baseline.

The performance of the 54 hyperparameter configurations on the evaluation dataset ranges from 0.34 to 0.47 in terms of AUC-PR. Among these configurations, 33 outperform the Sub-PCA baseline, 16 perform worse, and 5 achieve identical results.

\section{Discussion}
\label{sec:discussion}

This work shows that conclusions about reconstruction-based TSAD can change substantially depending on evaluation and pipeline choices that are often under-specified in the literature. While our experiments are restricted to univariate reconstruction methods under the offline TSB-AD protocol, several observations and practical recommendations emerge.

\paragraph{Inference windowing strategy is a critical component of reconstruction pipelines}
The inference procedure, in particular the stride and resulting overlap between reconstructed windows, has a critical and non-trivial impact on performance. Overlapping inference improves AUC-PR, AUC-ROC, and Standard-F1 for all evaluated models, with average relative gain up to +28\% and average absolute gain up to +0.10, and improvements on a majority of time series. This indicates that the detection quality of reconstruction-based methods depends not only on the architecture and training procedure, but also on how reconstructions are mapped back to point-wise anomaly scores.

\paragraph{Simple baselines remain competitive}
On TSB-AD, lightweight reconstruction models (AutoEncoder and DLinear) match or outperform more complex Transformer variants in the univariate setting. This reinforces the importance of including strong and carefully tuned baselines when claiming progress. In particular, architectural complexity alone does not guarantee improved anomaly detection performance, and tuning choices can lead to large performance gaps even for relatively simple models. Consequently, fair comparisons should report both the tuned configuration and the tuning budget (search space, selection metric, and number of seeds), as these can substantially influence the conclusions drawn from a benchmark. Additionally, \pca{}, the fastest method evaluated in this study, requires tuning only two parameters:  the subsequence length and the percentage of explained variance.  Yet, this method achieves strong results on the TSB-AD dataset, making it a compelling baseline method.

\paragraph{Benchmark heterogeneity and robustness matter}
Benchmark-level averages can hide substantial variability across datasets and time series instances. Our dataset-level analysis shows that some datasets yield high performance for most models, while other datasets remain challenging regardless of the architecture. In addition, deep models can exhibit non-negligible variability across random seeds, and performance can depend strongly on the selected hyperparameters. These observations support more robust reporting practices: (i) reporting results across multiple seeds (e.g., mean \(\pm\) std) rather than a single run; (ii) complementing benchmark-level averages with dataset-level breakdowns; and (iii) using statistical tests (e.g., paired tests across time series) to avoid over-interpreting small differences in averaged metrics. Overall, these practices help distinguish consistent improvements from gains that depend on favorable initialization, specific datasets, or particular hyperparameter choices.

\paragraph{Takeaway from hyperparameter tuning}
The retained hyperparameter configurations, shown in Table \ref{tab:model_search_space}, provide additional practical insight. The best configurations often use larger learning rates (frequently \(1\mathrm{e}{-2}\)) and longer training (often 50 epochs), and in several cases a constant learning rate performs as well as or better than \texttt{ReduceLROnPlateau}. This suggests that default settings commonly used in prior TSAD studies (e.g., smaller learning rates with aggressive scheduling) may be suboptimal for the univariate reconstruction setting considered here. More broadly, the observed sensitivity to hyperparameters motivates explicitly reporting the full training recipe (optimizer, learning rate, schedule, epochs, window length) and encourages future work on more robust selection criteria that reduce the dependence on labeled validation or extensive tuning.

\begin{table}[htb]
\centering
\caption{Hyperparameter selected configuration: learning rate (LR), number of training epochs (Ep.), subsequence length  (Len), learning-rate scheduler, and architecture preset (Arch).}
\label{tab:model_search_space}
\begin{tabular}{lccccc}
\toprule
\textbf{Model} & \textbf{LR} & \textbf{Ep.} & \textbf{Len} & \textbf{Sch.} & \textbf{Arch.} \\
\midrule
DLinear & $10^{-2}$ & 50 & 96 & Const & S \\
AutoEncoder & $10^{-2}$ & 50 & 96 & Plat. & S \\
FEDformer & $10^{-3}$ & 50 & 32 & Const & M \\
Transformer & $10^{-3}$ & 30 & 32 & Plat. & M \\
Autoformer & $10^{-4}$ & 50 & 32 & Const & L \\
TimesNet & $10^{-2}$ & 50 & 96 & Plat. & L \\
\bottomrule
\end{tabular}
\end{table}

\section{Conclusion}
\label{sec:conclusion}

We presented an empirical study of reconstruction-based univariate TSAD on the TSB-AD benchmark, complemented with an analysis on the UCR archive. Our results show that evaluation choices, and in particular the inference strategy (disjoint vs.\ overlapping windows), have a substantial and consistent impact on anomaly detection performance and can change method rankings. They also show that, under careful tuning, simple reconstruction baselines (e.g., AutoEncoder, DLinear) remain highly competitive, while performance varies widely across datasets and can be sensitive to random seeds and hyperparameters.

These findings support more transparent and reproducible evaluation of reconstruction-based TSAD: explicitly specifying inference settings (window length, stride, aggregation), reporting multi-seed results and dataset-level breakdowns, and using paired significance tests to avoid over-interpreting small average differences. 

A natural direction for future work is to extend this analysis to multivariate TSAD. We expect the same core issue to persist: multivariate reconstruction models still operate on sliding windows and must map multiple, potentially overlapping, window reconstructions back to point-wise scores. As a result, the choice of inference windowing strategy (stride, overlap, and aggregation) is likely to have a similar impact, potentially amplified by additional design choices specific to multivariate data (e.g., channel-wise vs.\ joint error aggregation, variable scaling, and correlation structure). Future studies should therefore evaluate whether overlap affects not only overall accuracy but also the localization of anomalous dimensions and the stability of rankings across datasets.

\section*{Acknowledgment}
This work is supported by Région Occitanie.
This work was performed using HPC resources from GENCI-IDRIS (Grant 2026-11017733).

\bibliographystyle{IEEEtran}
\bibliography{IEEEabrv,bibliography}

\begin{thebibliography}{10}
\providecommand{\url}[1]{#1}
\csname url@samestyle\endcsname
\providecommand{\newblock}{\relax}
\providecommand{\bibinfo}[2]{#2}
\providecommand{\BIBentrySTDinterwordspacing}{\spaceskip=0pt\relax}
\providecommand{\BIBentryALTinterwordstretchfactor}{4}
\providecommand{\BIBentryALTinterwordspacing}{\spaceskip=\fontdimen2\font plus
\BIBentryALTinterwordstretchfactor\fontdimen3\font minus
  \fontdimen4\font\relax}
\providecommand{\BIBforeignlanguage}[2]{{%
\expandafter\ifx\csname l@#1\endcsname\relax
\typeout{** WARNING: IEEEtran.bst: No hyphenation pattern has been}%
\typeout{** loaded for the language `#1'. Using the pattern for}%
\typeout{** the default language instead.}%
\else
\language=\csname l@#1\endcsname
\fi
#2}}
\providecommand{\BIBdecl}{\relax}
\BIBdecl

\bibitem{yanComprehensiveSurveyDeep2024}
P.~Yan, A.~Abdulkadir, P.-P. Luley, M.~Rosenthal, G.~A. Schatte, B.~F. Grewe,
  and T.~Stadelmann, ``A {{Comprehensive Survey}} of {{Deep Transfer Learning}}
  for {{Anomaly Detection}} in {{Industrial Time Series}}: {{Methods}},
  {{Applications}}, and {{Directions}},'' \emph{IEEE Access}, vol.~12, pp.
  3768--3789, 2024.

\bibitem{yangDetectionAnomalyStock2020}
W.~Yang, R.~Wang, and B.~Wang, ``Detection of {{Anomaly Stock Price Based}} on
  {{Time Series Deep Learning Models}},'' in \emph{2020 {{Management Science
  Informatization}} and {{Economic Innovation Development Conference}}
  ({{MSIEID}})}, 2020, pp. 110--114.

\bibitem{yangDeepLearningTechnologies2023}
X.~Yang, X.~Qi, and X.~Zhou, ``Deep {{Learning Technologies}} for {{Time Series
  Anomaly Detection}} in {{Healthcare}}: {{A Review}},'' \emph{IEEE Access},
  vol.~11, pp. 117\,788--117\,799, 2023.

\bibitem{zamanzadehdarbanDeepLearningTime2024}
\BIBentryALTinterwordspacing
Z.~Zamanzadeh~Darban, G.~I. Webb, S.~Pan, C.~Aggarwal, and M.~Salehi, ``Deep
  {{Learning}} for {{Time Series Anomaly Detection}}: {{A Survey}},'' vol.~57,
  no.~1, pp. 15:1--15:42. [Online]. Available:
  \url{https://dl.acm.org/doi/10.1145/3691338}
\BIBentrySTDinterwordspacing

\bibitem{bank2023autoencoders}
D.~Bank, N.~Koenigstein, and R.~Giryes, ``Autoencoders,'' \emph{Machine
  learning for data science handbook: data mining and knowledge discovery
  handbook}, pp. 353--374, 2023.

\bibitem{medsker2001recurrent}
L.~R. Medsker, L.~Jain \emph{et~al.}, ``Recurrent neural networks,''
  \emph{Design and applications}, vol.~5, no. 64-67, p.~2, 2001.

\bibitem{vaswaniAttentionAllYou2017}
\BIBentryALTinterwordspacing
A.~Vaswani, N.~Shazeer, N.~Parmar, J.~Uszkoreit, L.~Jones, A.~N. Gomez,
  L.~Kaiser, and I.~Polosukhin, ``Attention is {{All}} you {{Need}},'' in
  \emph{Advances in {{Neural Information Processing Systems}}}, vol.~30.\hskip
  1em plus 0.5em minus 0.4em\relax Curran Associates, Inc. [Online]. Available:
  \url{https://proceedings.neurips.cc/paper/2017/hash/3f5ee243547dee91fbd053c1c4a845aa-Abstract.html}
\BIBentrySTDinterwordspacing

\bibitem{qiuTABUnifiedBenchmarking2025}
\BIBentryALTinterwordspacing
X.~Qiu, Z.~Li, W.~Qiu, S.~Hu, L.~Zhou, X.~Wu, Z.~Li, C.~Guo, A.~Zhou, Z.~Sheng,
  J.~Hu, C.~S. Jensen, and B.~Yang, ``Tab: Unified benchmarking of time series
  anomaly detection methods,'' \emph{Proc. VLDB Endow.}, vol.~18, no.~9, p.
  2775–2789, May 2025. [Online]. Available:
  \url{https://doi.org/10.14778/3746405.3746407}
\BIBentrySTDinterwordspacing

\bibitem{liuElephantRoomReliable2025}
Q.~Liu and J.~Paparrizos, ``The elephant in the room: Towards a reliable
  time-series anomaly detection benchmark,'' in \emph{Proceedings of the 38th
  {{International Conference}} on {{Neural Information Processing Systems}}},
  ser. {{NIPS}} '24, vol.~37.\hskip 1em plus 0.5em minus 0.4em\relax Curran
  Associates Inc., 2024, pp. 108\,231--108\,261.

\bibitem{wangDeepTimeSeries2024}
\BIBentryALTinterwordspacing
Y.~Wang, H.~Wu, J.~Dong, Y.~Liu, M.~Long, and J.~Wang. Deep {{Time Series
  Models}}: {{A Comprehensive Survey}} and {{Benchmark}}. [Online]. Available:
  \url{http://arxiv.org/abs/2407.13278}
\BIBentrySTDinterwordspacing

\bibitem{chandola2009anomaly}
\BIBentryALTinterwordspacing
V.~Chandola, A.~Banerjee, and V.~Kumar, ``Anomaly detection: A survey,''
  \emph{ACM Comput. Surv.}, vol.~41, no.~3, Jul. 2009. [Online]. Available:
  \url{https://doi.org/10.1145/1541880.1541882}
\BIBentrySTDinterwordspacing

\bibitem{blazquez2021review}
\BIBentryALTinterwordspacing
A.~Bl\'{a}zquez-Garc\'{\i}a, A.~Conde, U.~Mori, and J.~A. Lozano, ``A review on
  outlier/anomaly detection in time series data,'' \emph{ACM Comput. Surv.},
  vol.~54, no.~3, Apr. 2021. [Online]. Available:
  \url{https://doi.org/10.1145/3444690}
\BIBentrySTDinterwordspacing

\bibitem{boniol2024dive}
P.~Boniol, Q.~Liu, M.~Huang, T.~Palpanas, and J.~Paparrizos, ``Dive into
  time-series anomaly detection: A decade review,'' 2024.

\bibitem{anomaly2023zhang}
A.~Zhang, S.~Deng, D.~Cui, Y.~Yuan, and G.~Wang, ``An experimental evaluation
  of anomaly detection in time series,'' \emph{PVLDB}, vol.~17, no.~3, 2023.

\bibitem{yeh2016matrix}
C.-C.~M. Yeh, Y.~Zhu, L.~Ulanova, N.~Begum, Y.~Ding, H.~A. Dau, D.~F. Silva,
  A.~Mueen, and E.~Keogh, ``Matrix profile i: all pairs similarity joins for
  time series: a unifying view that includes motifs, discords and shapelets,''
  in \emph{2016 IEEE 16th international conference on data mining
  (ICDM)}.\hskip 1em plus 0.5em minus 0.4em\relax Ieee, 2016, pp. 1317--1322.

\bibitem{liuIsolationForest2008}
F.~T. Liu, K.~M. Ting, and Z.-H. Zhou, ``Isolation {{Forest}},'' in
  \emph{ICDM}, 2008, pp. 413--422.

\bibitem{yaacobARIMABasedNetwork2010}
A.~H. Yaacob, I.~K. Tan, S.~F. Chien, and H.~K. Tan, ``{{ARIMA Based Network
  Anomaly Detection}},'' in \emph{2010 {{Second International Conference}} on
  {{Communication Software}} and {{Networks}}}, 2010, pp. 205--209.

\bibitem{10.1145/3219819.3219845}
K.~Hundman, V.~Constantinou, C.~Laporte, I.~Colwell, and T.~Soderstrom,
  ``Detecting spacecraft anomalies using lstms and nonparametric dynamic
  thresholding,'' in \emph{ACM SIGKDD}, New York, NY, USA, 2018, p. 387–395.

\bibitem{takeishi2019shapley}
N.~Takeishi, ``Shapley values of reconstruction errors of pca for explaining
  anomaly detection,'' in \emph{2019 international conference on data mining
  workshops (icdmw)}.\hskip 1em plus 0.5em minus 0.4em\relax IEEE, 2019, pp.
  793--798.

\bibitem{malhotraLSTMbasedEncoderDecoderMultisensor2016}
\BIBentryALTinterwordspacing
P.~Malhotra, A.~Ramakrishnan, G.~Anand, L.~Vig, P.~Agarwal, and G.~Shroff.
  {{LSTM-based Encoder-Decoder}} for {{Multi-sensor Anomaly Detection}}.
  arXiv.org. [Online]. Available: \url{https://arxiv.org/abs/1607.00148v2}
\BIBentrySTDinterwordspacing

\bibitem{parkMultimodalAnomalyDetector2018}
\BIBentryALTinterwordspacing
D.~Park, Y.~Hoshi, and C.~C. Kemp, ``A {{Multimodal Anomaly Detector}} for
  {{Robot-Assisted Feeding Using}} an {{LSTM-Based Variational Autoencoder}},''
  vol.~3, no.~3, pp. 1544--1551. [Online]. Available:
  \url{https://ieeexplore.ieee.org/abstract/document/8279425}
\BIBentrySTDinterwordspacing

\bibitem{zhangDeepNeuralNetwork2019}
\BIBentryALTinterwordspacing
C.~Zhang, D.~Song, Y.~Chen, X.~Feng, C.~Lumezanu, W.~Cheng, J.~Ni, B.~Zong,
  H.~Chen, and N.~V. Chawla, ``A {{Deep Neural Network}} for {{Unsupervised
  Anomaly Detection}} and {{Diagnosis}} in {{Multivariate Time Series Data}},''
  vol.~33, no.~01, pp. 1409--1416. [Online]. Available:
  \url{https://ojs.aaai.org/index.php/AAAI/article/view/3942}
\BIBentrySTDinterwordspacing

\bibitem{wu2022timesnet}
H.~Wu, T.~Hu, Y.~Liu, H.~Zhou, J.~Wang, and M.~Long, ``Timesnet: Temporal
  2d-variation modeling for general time series analysis,'' \emph{arXiv
  preprint arXiv:2210.02186}, 2022.

\bibitem{wenTransformersTimeSeries2023}
\BIBentryALTinterwordspacing
Q.~Wen, T.~Zhou, C.~Zhang, W.~Chen, Z.~Ma, J.~Yan, and L.~Sun. Transformers in
  {{Time Series}}: {{A Survey}}. [Online]. Available:
  \url{http://arxiv.org/abs/2202.07125}
\BIBentrySTDinterwordspacing

\bibitem{wu2021autoformer}
H.~Wu, J.~Xu, J.~Wang, and M.~Long, ``Autoformer: Decomposition transformers
  with auto-correlation for long-term series forecasting,'' \emph{Advances in
  neural information processing systems}, vol.~34, pp. 22\,419--22\,430, 2021.

\bibitem{zhou2022fedformer}
T.~Zhou, Z.~Ma, Q.~Wen, X.~Wang, L.~Sun, and R.~Jin, ``Fedformer: Frequency
  enhanced decomposed transformer for long-term series forecasting,'' in
  \emph{International conference on machine learning}.\hskip 1em plus 0.5em
  minus 0.4em\relax PMLR, 2022, pp. 27\,268--27\,286.

\bibitem{audibertUSADUnSupervisedAnomaly2020a}
\BIBentryALTinterwordspacing
J.~Audibert, P.~Michiardi, F.~Guyard, S.~Marti, and M.~A. Zuluaga, ``{{USAD}}:
  {{UnSupervised Anomaly Detection}} on {{Multivariate Time Series}},'' in
  \emph{Proceedings of the 26th {{ACM SIGKDD International Conference}} on
  {{Knowledge Discovery}} \& {{Data Mining}}}.\hskip 1em plus 0.5em minus
  0.4em\relax ACM, 2020, pp. 3395--3404. [Online]. Available:
  \url{https://dl.acm.org/doi/10.1145/3394486.3403392}
\BIBentrySTDinterwordspacing

\bibitem{xu2021anomaly}
J.~Xu, H.~Wu, J.~Wang, and M.~Long, ``Anomaly transformer: Time series anomaly
  detection with association discrepancy,'' \emph{arXiv preprint
  arXiv:2110.02642}, 2021.

\bibitem{kimContrastiveTimeSeriesAnomaly2024}
\BIBentryALTinterwordspacing
H.~Kim, S.~Kim, S.~Min, and B.~Lee, ``Contrastive {{Time-Series Anomaly
  Detection}},'' vol.~36, no.~10, pp. 5053--5065. [Online]. Available:
  \url{https://ieeexplore.ieee.org/abstract/document/10325644}
\BIBentrySTDinterwordspacing

\bibitem{fawcettIntroductionROCAnalysis2006}
\BIBentryALTinterwordspacing
T.~Fawcett, ``An introduction to {{ROC}} analysis,'' vol.~27, no.~8, pp.
  861--874. [Online]. Available:
  \url{https://www.sciencedirect.com/science/article/pii/S016786550500303X}
\BIBentrySTDinterwordspacing

\bibitem{davisRelationshipPrecisionRecallROC2006}
\BIBentryALTinterwordspacing
J.~Davis and M.~Goadrich, ``The relationship between {{Precision-Recall}} and
  {{ROC}} curves,'' in \emph{Proceedings of the 23rd International Conference
  on {{Machine}} Learning - {{ICML}} '06}.\hskip 1em plus 0.5em minus
  0.4em\relax ACM Press, pp. 233--240. [Online]. Available:
  \url{http://portal.acm.org/citation.cfm?doid=1143844.1143874}
\BIBentrySTDinterwordspacing

\bibitem{tatbulPrecisionRecallTime2018}
\BIBentryALTinterwordspacing
N.~Tatbul, T.~J. Lee, S.~Zdonik, M.~Alam, and J.~Gottschlich, ``Precision and
  {{Recall}} for {{Time Series}},'' in \emph{Advances in {{Neural Information
  Processing Systems}}}, vol.~31.\hskip 1em plus 0.5em minus 0.4em\relax Curran
  Associates, Inc. [Online]. Available:
  \url{https://proceedings.neurips.cc/paper/2018/hash/8f468c873a32bb0619eaeb2050ba45d1-Abstract.html}
\BIBentrySTDinterwordspacing

\bibitem{boniolVUSEffectiveEfficient2025}
\BIBentryALTinterwordspacing
P.~Boniol, A.~K. Krishna, M.~Bruel, Q.~Liu, M.~Huang, T.~Palpanas, R.~S. Tsay,
  A.~Elmore, M.~J. Franklin, and J.~Paparrizos, ``{{VUS}}: Effective and
  efficient accuracy measures for time-series anomaly detection,'' vol.~34,
  no.~3, p.~32. [Online]. Available:
  \url{https://doi.org/10.1007/s00778-025-00907-x}
\BIBentrySTDinterwordspacing

\bibitem{shyu2003novel}
M.-L. Shyu, S.-C. Chen, K.~Sarinnapakorn, and L.~Chang, ``A novel anomaly
  detection scheme based on principal component classifier,'' 2003.

\bibitem{zeng2023transformers}
A.~Zeng, M.~Chen, L.~Zhang, and Q.~Xu, ``Are transformers effective for time
  series forecasting?'' in \emph{Proceedings of the AAAI conference on
  artificial intelligence}, vol.~37, no.~9, 2023, pp. 11\,121--11\,128.

\bibitem{wilcoxon1992individual}
F.~Wilcoxon, ``Individual comparisons by ranking methods,'' in
  \emph{Breakthroughs in statistics: Methodology and distribution}.\hskip 1em
  plus 0.5em minus 0.4em\relax Springer, 1992, pp. 196--202.

\bibitem{holm1979simple}
S.~Holm, ``A simple sequentially rejective multiple test procedure,''
  \emph{Scandinavian journal of statistics}, pp. 65--70, 1979.

\bibitem{imamuraGeneralizedDiscordsTime2025a}
\BIBentryALTinterwordspacing
M.~Imamura, ``Generalized {{Discords}} for {{Time Series Anomaly Detection}}
  with {{Flexible Subsequence Lengths}},'' in \emph{Proceedings of the 31st
  {{ACM SIGKDD Conference}} on {{Knowledge Discovery}} and {{Data Mining
  V}}.2}, ser. {{KDD}} '25.\hskip 1em plus 0.5em minus 0.4em\relax Association
  for Computing Machinery, pp. 1013--1024. [Online]. Available:
  \url{https://dl.acm.org/doi/10.1145/3711896.3736977}
\BIBentrySTDinterwordspacing

\bibitem{tafazoliC22MPMarriageCatch222024}
\BIBentryALTinterwordspacing
S.~Tafazoli, Y.~Lu, R.~Wu, T.~V.~A. Srinivas, H.~Dela~Cruz, R.~Mercer, and
  E.~Keogh, ``{{C22MP}}: The marriage of catch22 and the matrix profile creates
  a fast, efficient and interpretable anomaly detector,'' \emph{Knowledge and
  Information Systems}, vol.~66, pp. 4789--4823, 2024. [Online]. Available:
  \url{https://doi.org/10.1007/s10115-024-02107-5}
\BIBentrySTDinterwordspacing

\bibitem{luDAMPAccurateTime2023}
\BIBentryALTinterwordspacing
Y.~Lu, R.~Wu, A.~Mueen, M.~A. Zuluaga, and E.~Keogh, ``{{DAMP}}: Accurate time
  series anomaly detection on trillions of datapoints and ultra-fast arriving
  data streams,'' \emph{Data Mining and Knowledge Discovery}, vol.~37, no.~2,
  pp. 627--669, 2023. [Online]. Available:
  \url{https://doi.org/10.1007/s10618-022-00911-7}
\BIBentrySTDinterwordspacing

\bibitem{boniolUnsupervisedScalableSubsequence2021}
\BIBentryALTinterwordspacing
P.~Boniol, M.~Linardi, F.~Roncallo, T.~Palpanas, M.~Meftah, and E.~Remy,
  ``Unsupervised and scalable subsequence anomaly detection in large data
  series,'' \emph{The VLDB Journal}, vol.~30, no.~6, pp. 909--931, 2021.
  [Online]. Available: \url{https://doi.org/10.1007/s00778-021-00655-8}
\BIBentrySTDinterwordspacing

\bibitem{zhuMatrixProfileXI2018}
\BIBentryALTinterwordspacing
Y.~Zhu, C.-C.~M. Yeh, Z.~Zimmerman, K.~Kamgar, and E.~Keogh, ``Matrix {{Profile
  XI}}: {{SCRIMP}}++: {{Time Series Motif Discovery}} at {{Interactive
  Speeds}},'' in \emph{2018 {{IEEE International Conference}} on {{Data
  Mining}} ({{ICDM}})}, 2018, pp. 837--846. [Online]. Available:
  \url{https://ieeexplore.ieee.org/abstract/document/8594908}
\BIBentrySTDinterwordspacing

\bibitem{tuliTranADDeepTransformer2022b}
\BIBentryALTinterwordspacing
S.~Tuli, G.~Casale, and N.~R. Jennings, ``{{TranAD}}: Deep transformer networks
  for anomaly detection in multivariate time series data,'' vol.~15, no.~6, pp.
  1201--1214. [Online]. Available:
  \url{https://dl.acm.org/doi/10.14778/3514061.3514067}
\BIBentrySTDinterwordspacing

\end{thebibliography}

\appendix

\begin{table}[h]
\centering
\caption{Training and Optimization Hyperparameter Search Space for the three random seeds $\{0, 1, 2\}$}
\label{tab:hypers}
\begin{tabular}{ll}
\toprule
Hyperparameter      & Search Values / Range                     \\ \midrule
Initial Learning Rate        & $\{10^{-2}, 10^{-3}, 10^{-4}\}$                    \\
Training Epochs              & $\{10, 20, 30, 50\}$                                \\
Subsequence Length           & $\{32, 64, 96\}$                                    \\
Learning Rate Scheduler      & Constant, \texttt{ReduceLROnPlateau}                \\
Scheduler Patience           & 5 epochs                                            \\
Scheduler Reduction Factor   & 0.5                                                 \\
Inference Strategy           & Disjoint, overlapping                               \\
\bottomrule
\end{tabular}
\end{table}

\begin{table*}[h]
\centering
\caption{Detailed Architecture Presets for Evaluated Models. Each model is evaluated across three capacity levels (Small, Medium, Large) to ensure a fair comparison of structural complexity. Bold indicates retained parameters.}
\label{tab:model_specific_presets}
\small
\begin{tabular}{llccc}
\toprule
Model & Architectural Parameter & Small & Medium & Large \\ 
\midrule

\multirow{1}{*}{\pca{}} 
& Percentage of explained variance & 25 & 50 & \bf75 \\
\midrule

\multirow{1}{*}{DLinear} 
& Moving average ratio & \bf0.1 & 0.25 & 0.5 \\
\midrule

\multirow{1}{*}{AutoEncoder} 
& Hidden dimensions ratio & \bf[0.5, 0.25] & [0.75, 0.5, 0.25] & [1, 0.75, 0.5, 0.25] \\
\midrule

\multirow{4}{*}{Transformer} 
& Model dimension ($d_{model}$) & 8 & \bf16 & 32 \\
& Feed-forward dimension ($d_{ff}$) & 16 & \bf32 & 64 \\
& Encoder layers ($e_{layers}$) & 1 & \bf2 & 3 \\
& Attention heads ($n_{heads}$) & 2 & \bf2 & 4 \\
\midrule

\multirow{5}{*}{FEDformer} 
& Model dimension ($d_{model}$) & 8 & \bf16 & 32 \\
& Feed-forward dimension ($d_{ff}$) & 16 & \bf32 & 64 \\
& Encoder layers ($e_{layers}$) & 1 & \bf1 & 2 \\
& Attention heads ($n_{heads}$) & 2 & \bf2 & 4 \\
& Moving average window & 15 & \bf25 & 25 \\
\midrule

\multirow{5}{*}{Autoformer} 
& Model dimension ($d_{model}$) & 8 & 16 & \bf32 \\
& Feed-forward dimension ($d_{ff}$) & 16 & 32 & \bf64 \\
& Encoder layers ($e_{layers}$) & 1 & 1 & \bf2 \\
& Attention heads ($n_{heads}$) & 2 & 2 & \bf4 \\
& Factor & 3 & 3 & \bf5 \\
& Moving average window & 15 & 25 & \bf25 \\
\midrule

\multirow{5}{*}{TimesNet} 
& Top-k & 3 & 3 & \bf3 \\
& Model dimension ($d_{model}$) & 8 & 12 & \bf16 \\
& Feed-forward dimension ($d_{ff}$) & 16 & 24 & \bf32 \\
& Number of kernels & 3 & 4 & \bf6 \\
& Encoder layers ($e_{layers}$) & 1 & 1 & \bf2 \\
\bottomrule

\end{tabular}
\end{table*}

\end{document}